# Possibilistic Conditioning and Propagation

**Yen-Teh Hsia**
Department of Information and Computer Engineering
Chung Yuan Christian University
Chung-Li, Taiwan 32023, R.O.C.

## Abstract

We give an axiomatization of confidence transfer - a known conditioning scheme - from the perspective of expectation-based inference in the sense of Gärdenfors and Makinson. Then, we use the notion of belief independence to "filter out" different proposals of possibilistic conditioning rules, all are variations of confidence transfer. Among the three rules that we consider, only Dempster's rule of conditioning passes the test of supporting the notion of belief independence. With the use of this conditioning rule, we then show that we can use local computation for computing desired conditional marginal possibilities of the joint possibility satisfying the given constraints. It turns out that our local computation scheme is already proposed by Shenoy. However, our intuitions are completely different from that of Shenoy. While Shenoy just defines a local computation scheme that fits his framework of valuation-based systems, we derive that local computation scheme from $\Pi(\beta) = \Pi(\beta \mid \alpha) * \Pi(\alpha)$ and appropriate independence assumptions, just like how the Bayesians derive their local computation scheme.

## 1 INTRODUCTION

The theory of possibility [Zadeh 78] formalizes a notion of belief that may be described as follows. Let $\mathcal{P}$ be a finite, non-empty set of propositional primitives. Let $\mathcal{L}_\mathcal{P}$ be the least set of formulas containing $\mathcal{P}$, closed under $\neg$ and $\wedge$ (with the usual abbreviations $\vee$ and $\supset$). Let $\Theta$ be the set of all interpretations of $\mathcal{P}$. A *possibility measure* $\Pi$ is a function from the set of all propositions $2^\Theta$ to the real interval $[0, 1]$. $\Pi$ satisfies three conditions:[1]

(i) $\Pi(\bot) = 0$ /* it is impossible for any unsatisfiable proposition to be true */

(ii) $\Pi(\top) = 1$ /* a tautology is never less than totally possible (i.e., always 1) */

(iii) $\Pi(\alpha \vee \beta) = \max(\Pi(\alpha), \Pi(\beta))$
/* the basic postulate of possibility */

For any proposition $\alpha$, $1 - \Pi(\alpha)$ represents "*my potential degree of surprise* upon realizing that $\alpha$ is true" [Shackle 61; Hsia 91]. $\Pi$ is such that either $\Pi(\alpha) = 1$, or $\Pi(\neg \alpha) = 1$, or both, since $\max(\Pi(\alpha), \Pi(\neg \alpha)) = \Pi(\alpha \vee \neg \alpha) = \Pi(\top) = 1$. Therefore by equating "believe $\alpha$" with $\Pi(\neg \alpha) < 1$ (meaning that $\neg \alpha$ is considered to be less possible, while $\alpha$ is considered to be totally possible), we see that the notion of belief here is that we do not believe in both $\alpha$ and $\neg \alpha$ at the same time. This conforms to our everyday use of the word 'believe'.

From $\Pi$, we can define its *density function* $\pi : \Theta \to [0, 1]$, where $\pi(\omega) = \Pi(\{\omega\})$. $\pi$ is also known as a *possibility distribution*. It completely determines $\Pi$. And so one other way of understanding possibility theory is to postulate the existence of a possibility distribution $\pi$ on $\Theta$. $\pi$ is such that there is always a world $\omega$ that is totally possible ($\pi(\omega) = 1$), while $\Pi$ is induced by $\pi$ in that $\Pi(A) = \max\{\pi(\omega): \omega \in A\}$. This view of possibility theory has a close analogy with probability theory (just replace maximum by addition) : a probability distribution p is such that $\sum\{p(\omega) : \omega \in \Theta\} = 1$; moreover, $P(A) = \sum\{p(\omega) : \omega \in A\}$.

Over the years, there have been several suggestions concerning the "right" conditioning rule(s) to use in possibility theory. Two often-mentioned rules are, respectively, the minimum rule of conditioning (what Dubois and Prade [86, 91] use in their reasoning approach) and Dempster's rule of conditioning (the "official" conditioning rule that is used in the theory of belief function [Shafer 76; Smets 88]). These two rules, together with a third but less-known rule due to Yager [87], are all instances of a generic conditioning scheme called *confidence transfer* (also known as *belief transfer* [Smets 88]). And so two questions are of interest. One, is there any reason why we should be interested in confidence transfer? Two, if we are to use one particular instance of confidence transfer as our "official" conditioning rule, which conditioning rule should we use?

---

[1] We write $\alpha$ to mean either a formula $\alpha$ or a set $[\alpha]$ ($\subseteq \Theta$); $[\alpha]$ is the set of all models of $\alpha$.

In this paper, we seek to justify confidence transfer from a logic-based perspective. Specifically, we give an axiomatization from the standpoint of expectation-based inference in the sense of Gärdenfors and Makinson [94]. By strengthening a proposal for making belief revisions that is due to Rott [91], we derive the notion of confidence transfer. To choose an appropriate instance of confidence transfer as our official rule of conditioning, we then use the notion of belief independence to "filter out" different proposals. Among the three rules of conditioning that we mentioned, only Dempster's rule passes the test of supporting the notion of belief independence. Though we cannot rule out all possible alternatives (there are infinitely many of them), Dempster's rule does seem to be our only natural choice.

Other authors have also tried to justify some of the conditioning rules that we mentioned. Dubois and Prade [86], for example, used an equation ((I) below) proposed by Hisdal [78] to justify both the minimum rule and Dempster's rule.[2]

$$\Pi(\alpha \wedge \beta) = \Pi(\beta \mid \alpha) \text{ op } \Pi(\alpha), \text{ where op is a binary operator} \quad (I)$$

Smets [93], on the other hand, gives an explicit axiomatization of Dempster's rule of conditioning in a rather abstract setting (where he talks about *all* "reasonably defined" credibility functions). It is interesting to note that both of these two attempts are related to the Bayesian approach in some way. Equation (I), for example, originates directly from the probabilistic conditioning rule $P(\alpha \wedge \beta) = P(\beta \mid \alpha) * P(\alpha)$ [Hisdal 78]. And even though Smets works on an abstract level, he nevertheless requires that *if* probabilities are used as credibility functions, then the general conditioning rule that he is trying to derive *must* reduce to a rule that preserves additivity. Thus, these authors more or less rely on the a priori existence of the Bayesian approach in order for their work to be fully meaningful. This is not the case here. What we need here is an a priori notion of belief independence, among others.

Suppose we do use Dempster's rule of conditioning as our official rule of conditioning, we can then use the so-called "causal net" technique [Pearl 88] for modular specifications, and use local computation for computing conditional marginal possibilities. As it turns out, our local computation scheme is already described in Shenoy [92]. However, the underlying intuitions are completely different. Shenoy [89] builds his framework of valuation-based systems on top of the three axioms of local computation proposed by Shenoy and Shafer [90]. Then he just defines a local computation scheme for possibility theory that fits in his framework. Here, we *derive* that local computation scheme from $\Pi(\beta) = \Pi(\beta \mid \alpha) * \Pi(\alpha)$ and appropriate independence assumptions, just like how the Bayesians derive their local computation scheme (see, for example, [Pearl 86; Lauritzen and Spiegelhalter 88]).

There are also other related work in the past [Dubois and Prade 90a; Fonck 90]. But the aims and tasks are completely different. Dubois and Prade [90a], for example, base their work on the assumption that possibilities can be combined, and they showed that their combination scheme can be computed using local computation. Fonck [90] also reported what amounts to a special case of the work of Dubois and Prade [90a]. The main difference between these past works and ours is in the kind of problems solved. For Dubois and Prade [90a] and Fonck [90], the problem is to compute (using local computation) marginalizations of a *combined* possibility measure (on $2^\Theta$) from several given possibility measures (also on $2^\Theta$). For us, the problem here is to compute (again using local computation) marginalizations of the *joint* possibility measure satisfying the given constraints.

The rest of this paper is organized as follows. In Section 2, we introduce the inference paradigm of Gärdenfors and Makinson [94], and we show how confidence transfer can be derived by strengthening their notion of inference in a perfectly reasonable way. In Section 3, we introduce the notion of belief independence, and show that Dempster's rule is the only one that is acceptable among the three rules that we consider. In Section 4, we describe how we can use the proposal of Pearl [86] to specify "prior" possibilities and conditional possibilities, and we use local computation to compute desired marginals from the given specifications. Section 5 concludes.

## 2 AN AXIOMATIC JUSTIFICATION OF CONFIDENCE TRANSFER

Recently, Gärdenfors and Makinson [94] proposed the following approach for nonmonotonic reasoning. First, we give two sets $\Gamma$ and $\Delta$ of formulas; $\Gamma$ is a set of hard, non-defeasible constraints (e.g., $\Gamma$ = {Penguin $\supset$ Bird}); $\Delta$ includes $\Gamma$ and all of our soft, defeasible expectations (e.g., $\Delta = \Gamma \cup$ {Bird $\supset$ Fly, Penguin $\supset \neg$ Fly, ...}),[3] and is closed under logical consequence. Next, we assume the existence of an ordering $\geq_E$ of *all* sentences. $\geq_E$ is called an *expectation ordering*, and satisfies the following axioms. (Below, $\alpha$, $\beta$, and $\gamma$ are propositional formulas, while $\vdash$ is the usual propositional provability symbol.)

(E1) If $\alpha \geq_E \beta$ and $\beta \geq_E \gamma$, then $\alpha \geq_E \gamma$;

(E2) If $\Gamma \vdash \alpha \supset \beta$, then $\beta \geq_E \alpha$;

(E3) Either $\alpha \wedge \beta \geq_E \alpha$ or $\alpha \wedge \beta \geq_E \beta$;

---

[2] Dubois and Prade [86, 88] also used what they call *the principle of minimum specificity* in their justification.

[3] We are assuming that some extraordinary penguins do fly.

$\alpha \geq_E \beta$ means $\alpha$ is *at least as expected as* $\beta$ (or $\neg \alpha$ is *at least as surprising as* $\neg \beta$). We write $\alpha >_E \beta$ to mean that $\alpha \geq_E \beta$ and not $\beta \geq_E \alpha$, and we write $\alpha =_E \beta$ to mean that $\alpha \geq_E \beta$ and $\beta \geq_E \alpha$.

Once we have specified $\Gamma$ and $\Delta$, and also postulated the existence of $\leq_E$, we can then make inferences from our given information $\alpha$ according to the following principle of reasoning that is due to Rott [91].

> $\alpha$ nonmonotonically entails $\gamma$, denoted as $\alpha \mid\sim_E \gamma$, if and only if
> $\Gamma \cup \{\alpha\} \cup \{\beta : \beta \in \Delta \text{ and } \beta >_E \neg \alpha\} \mid\!- \gamma$.

$\mid\sim_E$ is called a *comparative expectation inference relation*.

As it happens, (E1) - (E3), together with a fourth axiom (call it E4) that $\top > \bot$, are equivalent to the axioms underlying necessity measures [Dubois 86; Dubois and Prade 90b] - the mathematical equivalent of consonant belief functions [Shafer 76].

A *consonant m-value function* (or just *m-value function*) is a function m: $2^\Theta \to [0, 1]$ that satisfies the following three conditions:

(1) $\Sigma_{B \subseteq \Theta} m(B) = 1$,  /* confidence in $\top$ must be the maximum possible */

(2) $m(\varnothing) = 0$, and  /* there must be no confidence in $\bot$ */

(3) There are nested subsets of $\Theta$ $S_1, S_2, ..., S_N$ ($S_1 \subset S_2 \subset ... \subset S_N \subseteq \Theta$) such that $m(x) \neq 0$ if and only if $x \in \{S_1, S_2, ..., S_N\}$. ($S_1, S_2, ..., S_N$ are called the *focal elements* of m.)  /* confidence is structured hierarchically */

Every consonant m-value function induces a *consonant belief function*
$Co_m: 2^\Theta \to [0, 1]$ as follows: $Co_m(\alpha) = \Sigma_{B \subseteq [\alpha]} m(B)$, that is, the $Co_m$ value of a formula $\alpha$ is computed by adding up the m-values of all subsets of $[\alpha]$. (Below, we write Co instead of $Co_m$ whenever no confusion will result.)

Co is the mathematical dual of $\Pi$, and the relation is that $Co(\alpha) = 1 - \Pi(\neg \alpha)$.

Since (E1) - (E4) actually characterize all consonant belief functions, it is tempting to think that perhaps we can view the comparative expectation inference relation of Gärdenfors and Makinson [94] and Rott [91] as some kind of updating (or conditioning) mechanism for consonant belief functions. But to translate the proposal of Rott into the framework of consonant belief functions, we need to solve two problems. The first problem is as follows. Even though it is assumed that there is a complete ordering $\geq_E$ on $\mathcal{L}_P$ when there is no information (i.e., $\alpha$ is just $\top$), nothing is said about the (new) ordering on $\mathcal{L}_P$ when there *is* some information. To be fair, this is not the concern of either Rott [91] or Gärdenfors and Makinson [94], for they are only interested in what *ought to be inferred* when the given information is $\alpha$. But here we obviously need to say something meaningful about this new ordering. As for the second "problem" with Rott's original proposal, it is just that quantification is not considered for obvious reasons. Below, we strengthen Rott's proposal so that both problems are solved in a reasonable way. We need a basic axiom though.

(**D1**) $Co(.|\alpha)$ is a consonant belief function on $\Theta$.

Intuitively, D1 just says that we *still* have a complete ordering on $\mathcal{L}_P$ when we are given the information $\alpha$; this ordering satisfies (E1) - (E4), and $Co(.|\alpha)$ is its numerical counterpart. (Below, we use $Co(\beta \mid \alpha)$ as a shorthand for $Co(.|\alpha)(\beta)$.)

Let us now see how a new ordering can be inferred from the old ordering and $\alpha$. Following Gärdenfors and Makinson [94], we let $\Delta$ be the set $\{\beta : \beta >_E \bot\}$. This shortens the definition of $\alpha \mid\sim_E \gamma$ to $\Gamma \cup \{\alpha\} \cup \{\beta : \beta >_E \neg \alpha\} \mid\!- \gamma$. When $\Gamma \cup \{\alpha\}$ is consistent, we arrive at an even simpler definition: $\{\alpha\} \cup \{\beta : \beta >_E \neg \alpha\} \mid\!- \gamma$. Translated into consonant belief functions, $\alpha \mid\sim_E \gamma$ becomes $Co(\gamma \mid \alpha) > 0$, and Rott's proposal becomes the following.

$$Co(\gamma \mid \alpha) > 0 \text{ iff } \exists \beta, Co(\beta) > Co(\neg \alpha) \text{ and} \\ \mid\!- \alpha \wedge \beta \supset \gamma. \quad \text{(II)}$$

We remark that (II) actually should be written as "if $Co(\neg \alpha) < 1$ then (II)", since the prerequisite for (II) is that $\Gamma \cup \{\alpha\}$ is consistent (which is translated into $Co(\neg \alpha) < 1$). However, our axiomatization will be such that $Co(.|\alpha)$ is undefined when $Co(\neg \alpha) = 1$. And so to facilitate our discussions, we leave (II) as it is. Lemma 2.1 below shows that (III) is a reformulation of (II).

$$Co(\gamma \mid \alpha) > 0 \text{ iff } Co(\alpha \supset \gamma) > Co(\neg \alpha). \quad \text{(III)}$$

**Lemma 2.1.** $Co(\alpha \supset \gamma) > Co(\neg \alpha)$ iff $\exists \beta$, $Co(\beta) > Co(\neg \alpha)$ and $\mid\!- \alpha \wedge \beta \supset \gamma$.

**Proof:** ($\Rightarrow$) just let $\beta$ be $\alpha \supset \gamma$.
($\Leftarrow$) Let $\beta$ be such that $Co(\beta) > Co(\neg \alpha)$ and $\mid\!- \alpha \wedge \beta \supset \gamma$. $\mid\!- \beta \supset (\alpha \supset \gamma)$. And so $Co(\alpha \supset \gamma) \geq Co(\beta) > Co(\neg \alpha)$.  □

Thus by Rott's proposal, to see whether $\gamma$ ought to be expected when the given information is $\alpha$, we just see if our (original) expectation of $\alpha \supset \gamma$ is higher than our expectation of $\neg \alpha$. But still, nothing is said about the new ordering among those newly expected propositions. In particular, how should we order $\beta$ and $\gamma$ if $Co(\alpha \supset \beta) = Co(\alpha \supset \gamma)$, and how should we order $\beta$ and $\gamma$ if $Co(\alpha \supset \beta) > Co(\alpha \supset \gamma)$? Below, (D2) states that when the context is $\alpha$, the difference in our (new) expectations of $\beta$ and $\gamma$ should at least match the difference in our (original) expectations of $\alpha \supset \beta$ and $\alpha \supset \gamma$.

(**D2**) If $Co(\alpha \supset \beta) \geq Co(\alpha \supset \gamma)$
then $Co(\beta \mid \alpha) - Co(\gamma \mid \alpha) \geq Co(\alpha \supset \beta) - Co(\alpha \supset \gamma)$

(D2) implies that if $\beta$ and $\gamma$ are such that $\alpha \supset \beta$ and $\alpha \supset \gamma$ are equally expected in the first place, then our expectations of $\beta$ and $\gamma$ should rationally be the same when the context is that $\alpha$ is true. Thus, $Co(\alpha \mid \alpha) = Co(T \mid \alpha) = 1$, since $Co(\alpha \supset \alpha) = Co(\alpha \supset T)$. This means $Co(\neg \alpha \mid \alpha) = 0$ (by D1). For any $\gamma$, either $Co(\alpha \supset \gamma) > Co(\alpha \supset \neg \alpha)$ or $Co(\alpha \supset \gamma) = Co(\alpha \supset \neg \alpha)$ (since $\vdash \neg \alpha \supset (\alpha \supset \gamma)$). In the first case, $Co(\gamma \mid \alpha) - Co(\neg \alpha \mid \alpha) \geq Co(\alpha \supset \gamma) - Co(\alpha \supset \neg \alpha) > 0$, i.e., $Co(\gamma \mid \alpha) > 0$. In the second case, $Co(\gamma \mid \alpha) = Co(\neg \alpha \mid \alpha) = 0$. And so (III), i.e., Rott's proposal, is a logical consequence of (D2).

(D2) only defines the relative differences among the new expectations. Below, (D3) states that *if* we ever want to decrease our expectation of $\beta$ when the context is $\alpha$, we should not be overdoing it. After all, it is $\neg \alpha$ that we use as the "threshold" for determining new expectations in the first place. And so the decrease should not be more than $Co(\neg \alpha)$, our original expectation of $\neg \alpha$.

(**D3**) $Co(\beta \mid \alpha) \geq Co(\beta) - Co(\neg \alpha)$

This completes our axiomatization of conditioning. Note that if $Co(\neg \alpha)$ is 1, then $Co(.\mid\alpha)$ is undefined due to (D1) and (D2). We remark that Dubois and Prade [91, Proposition 5] proved something similar to (III). But what they did is completely different from what we are doing here. While Dubois and Prade [91] derived something similar to (III) from some *given* definition of $Co(.\mid\alpha)$ (i.e., their version of the minimum rule), we are trying to strengthen (III) and then *derive* the definition (a family of definitions in fact) of $Co(.\mid\alpha)$.

Suppose $Co(\neg \alpha) < 1$. Let us now find an algorithm for deriving $Co(.\mid\alpha)$ from $Co$ and $\alpha$. From (D1), we know we must construct a consonant belief function on $\Theta$. (D2) implies that $Co(\alpha\mid\alpha) = 1$. Therefore all focal elements of $Co(.\mid\alpha)$ are subsets of $[\alpha]$. Let the focal elements of $Co(.\mid\alpha)$ be $S_1, S_2, ..., S_n$ ($S_1 \subset S_2 \subset ... \subset S_n \subseteq [\alpha]$). Let $R_1, R_2, ..., R_m$ be the focal elements of $Co$ such that $R_1 \subset R_2 \subset ... \subset R_m \subseteq \Theta$ and $R_1$ is the smallest focal element of $Co$ that has a non-empty intersection with $[\alpha]$ (the existence of $R_1$ is guaranteed by $Co(\neg \alpha) < 1$). (D2) implies that if $Co(\alpha \supset \beta) = Co(\alpha \supset \gamma)$, then $Co(\beta \mid \alpha) = Co(\gamma \mid \alpha)$. And so for every $S_j$, there is an $R_i$ such that $S_j = R_i \cap [\alpha]$. (D2) also implies that if $Co(\alpha \supset \beta) > Co(\alpha \supset \gamma)$, then $Co(\beta \mid \alpha) - Co(\gamma \mid \alpha)$ ($\geq Co(\alpha \supset \beta) - Co(\alpha \supset \gamma)) > 0$. And so for every $R_i$, there is an $S_j$ such that $S_j = R_i \cap [\alpha]$. For every $j$ ($1 \leq j \leq n$), define $t(j)$ as follows: $R_1, R_2, ..., R_{t(1)}$ are all those $R_i$'s such that $S_1 = R_i \cap [\alpha]$, and $R_{t(j-1)+1}, R_{t(j-1)+2}, ..., R_{t(j)}$ are all those $R_i$'s such that $S_j = R_i \cap [\alpha]$. By (D3), we know that $m(S_1 \mid \alpha) = Co(S_1 \mid \alpha) = Co([\neg \alpha] \cup S_1 \mid \alpha) \geq Co([\neg \alpha] \cup S_1) - Co(\neg \alpha) = m(R_1) + m(R_2) + ... + m(R_{t(1)})$ ($m(S_1 \mid \alpha)$ is a shorthand for $m(.\mid\alpha)(S_1)$, and $m(.\mid\alpha)$ is the m-value function associated with $Co(.\mid\alpha)$). By (D2), we know that for every $j$ ($2 \leq j \leq n$), $m(S_j \mid \alpha) = Co(S_j \mid \alpha) - Co(S_{j-1} \mid \alpha) \geq Co([\neg \alpha] \cup S_j) - Co([\neg \alpha] \cup S_{j-1}) = m(R_{t(j-1)+1}) + m(R_{t(j-1)+2}) + ... + m(R_{t(j)})$. Let $m(S_1 \mid \alpha) = m(R_1) + m(R_2) + ... + m(R_{t(1)}) + c_1$; and for every $j$ ($2 \leq j \leq n$), let $m(S_j \mid \alpha) = m(R_{t(j-1)+1}) + m(R_{t(j-1)+2}) + ... + m(R_{t(j)}) + c_j$, where $c_j$'s are non-negative numbers. Then $\sum c_j = \sum_{j=1,n} m(S_j \mid \alpha) - \sum_{i=1,m} m(R_i) = 1 - \sum_{j=1,m} m(R_i) = Co(\neg \alpha)$. This is a known conditioning scheme called *confidence transfer* (also known as *belief transfer* [Smets 88]).[4] Below, we give its definition in terms of possibilities so as to facilitate our discussion in the next sections.

**Confidence transfer (the algorithm).** Let $\Pi$ be a possibility measure over $2^\Theta$. If $\Pi(\alpha) \neq 0$ (i.e., $\alpha$ is not considered impossible), then $\pi(.\mid\alpha)$, the density function of $\Pi(.\mid\alpha)$, can be computed using the following algorithm.

(1) If $\omega \notin [\alpha]$, then $\pi(\omega\mid\alpha)$ is assigned 0.

(2) As for elements of $[\alpha]$, do the following.

First partition the elements of $[\alpha]$ by their (original) possibilities, that is,

let $\{A_1, A_2, ..., A_n\}$ be a partition of $[\alpha]$ such that $\forall i, j \in \{1, 2, ..., n\}$, $\forall \omega_i \in A_i, \forall \omega_j \in A_j$, $\pi(\omega_i) = \pi(\omega_j)$ if $i = j$, and $\pi(\omega_i) < \pi(\omega_j)$ if $i < j$.

Next, choose $n$ non-negative real numbers $c_1, c_2, ..., c_n$ according to some arbitrary but fixed rule, so that $\Sigma_{i=1,n} c_i = 1 - \Pi(\alpha)$. (We call this rule a *normalization rule* and $1 - \Pi(\alpha)$ *the normalization constant*).

Finally, $\forall \omega \in A_i$, $\pi(\omega\mid\alpha)$ is assigned the value $\pi(\omega) + c_1 + c_2 + ... + c_i$.

Note that in the algorithm of confidence transfer, there is a parameter - a normalization rule - that still needs to be specified. Depending on what this rule is, different conditioning rules can result. Theorem 2.2 summarizes the above discussion.

**Theorem 2.2. (Axiomatic justification of confidence transfer)**

Let $\geq_E$ be an expectation ordering on $\mathcal{L}_P$, and let $\Pi$ be an element of the family of possibility measures induced by $\geq_E$ ($\Pi(\neg\alpha) \leq \Pi(\neg\beta)$ iff $\alpha \geq_E \beta$).

---

[4] The name "confidence transfer" is used here, because essentially what happens is that the (old) m-value of a set $[\alpha \supset \beta]$ is "transferred" to the set $[\beta]$ (and we just add up what $[\beta]$ receives).

Suppose $\Pi(\alpha) \neq 0$. Then the set of possibility measures $\Pi(.|\alpha)$ defined by (D1) - (D3) is the same as the set of possibility measures that are obtained from $\Pi$ and $\alpha$ using the algorithm of confidence transfer.

## 3 BELIEF INDEPENDENCE AND DEMPSTER'S RULE OF CONDITIONING

Consider again the algorithm of confidence transfer described at the end of the last section. Suppose we use a normalization rule that assigns $1 - \Pi(\alpha)$ to $c_1$ and 0's to $c_2, ..., $ and $c_n$. Then this is the conditioning rule that Yager [87] proposed. If, on the other hand, we let $c_n$ be $1 - \Pi(\alpha)$ and $c_1, ..., $ and $c_{n-1}$ be 0, then this is essentially the minimum rule of conditioning that Dubois and Prade [86, 91] use in their reasoning approach.[5] But if we distribute $1 - \Pi(\alpha)$ in such a way that $\pi(\omega|\alpha) = \pi(\omega)/\Pi(\alpha)$, then we will be working with Dempster's rule of conditioning. With this rule, $\Pi(\beta|\alpha) = \Pi(\alpha \wedge \beta|\alpha) = \max\{\pi(\omega|\alpha) : \omega \in [\alpha \wedge \beta]\} = \max\{\pi(\omega)/\Pi(\alpha) : \omega \in [\alpha \wedge \beta]\} = \Pi(\alpha \wedge \beta) / \Pi(\alpha)$. This looks just like the probabilistic rule of conditioning, with P (probability) replaced by $\Pi$ (possibility).

In order to choose among all these different proposals for the normalization rule, we use the notion of belief independence. The idea is as follows. Suppose $\alpha$ and $\beta$ are two propositions that are judged to be independent of each other. Then whatever we learn about $\alpha$ (alternatively, $\beta$), it should not have any effect on our opinion about $\beta$ (alternatively, $\alpha$). Any conditioning rule that we use should allow us to express this intuition. This means, on the formal level, that there must exist a solution, a possibility measure, that satisfies the following constraints when none of $\Pi(\alpha)$, $\Pi(\neg\alpha)$, $\Pi(\beta)$, and $\Pi(\neg\beta)$ is 0.

$\Pi(\alpha) = \Pi(\alpha | \beta) = \Pi(\alpha | \neg\beta)$, $\Pi(\neg\alpha) = \Pi(\neg\alpha | \beta) = \Pi(\neg\alpha | \neg\beta)$,

$\Pi(\beta) = \Pi(\beta | \alpha) = \Pi(\beta | \neg\alpha)$, $\Pi(\neg\beta) = \Pi(\neg\beta | \alpha) = \Pi(\neg\beta | \neg\alpha)$.

To facilitate the proofs, let us consider the case in which both $\alpha$ and $\beta$ are expected; none of the expectations are total, and the expectations are different. In other words,

$0 < \Pi(\neg\beta) < \Pi(\neg\alpha) < 0.5$.

Now, if our conditioning rule is the minimum rule of conditioning, then there is no possibility measure $\Pi$ that satisfies the above constraints. Let us prove by contradiction. Suppose there is such a measure $\Pi$. Then $\Pi(\neg\alpha | \neg\beta) = \Pi(\neg\alpha) \neq 1$ (by the constraints). But $\Pi(\neg\alpha | \neg\beta) = \Pi(\neg\alpha \wedge \neg\beta | \neg\beta)$. And so $\Pi(\neg\alpha | \neg\beta) = \Pi(\neg\alpha \wedge \neg\beta | \neg\beta) = \Pi(\neg\alpha \wedge \neg\beta)$ (since the minimum rule is such that $\Pi(\neg\alpha \wedge \neg\beta | \neg\beta)$ is either $\Pi(\neg\alpha \wedge \neg\beta)$ or 1). This implies that $\Pi(\neg\alpha) = \Pi(\neg\alpha \wedge \neg\beta)$. And so $\Pi(\neg\beta) = \Pi(\neg\beta \vee (\neg\alpha \wedge \neg\beta)) = \max(\Pi(\neg\beta), \Pi(\neg\alpha \wedge \neg\beta)) = \max(\Pi(\neg\beta), \Pi(\neg\alpha))$, which contradicts our assumption that $\Pi(\neg\beta) < \Pi(\neg\alpha)$.

Similarly, if our conditioning rule is what Yager [87] proposed, then there is no possibility measure $\Pi$ that satisfies the above constraints either. Again we prove by contradiction. Suppose there is such a measure $\Pi$. Then $\Pi(\neg\alpha | \neg\beta) = \Pi(\neg\alpha \wedge \neg\beta | \neg\beta) = \Pi(\neg\alpha \wedge \neg\beta) + (1 - \Pi(\neg\beta))$ (since Yager's proposal is such that $\Pi(\gamma | \neg\beta) = \Pi(\gamma) + (1 - \Pi(\neg\beta))$ for every $\gamma$ such that $\gamma \supset \neg\beta$). And so $\Pi(\neg\alpha) = \Pi(\neg\alpha | \neg\beta) = \Pi(\neg\alpha \wedge \neg\beta) + (1 - \Pi(\neg\beta)) > 0.5$, which contradicts our assumption that $\Pi(\neg\alpha) < 0.5$.

Contrasting to these two rules, a unique solution exists if the rule we use is Dempster's rule of conditioning.

### Theorem 3.1. (Joint possibility from independent marginal possibilities)

Let $\alpha_1, \alpha_2, ..., \alpha_n$ be n mutually exclusive formulas (i.e., exactly one of the $\alpha_i$'s is true), and let $\beta_1, \beta_2, ..., \beta_m$ be m mutually exclusive formulas. Suppose that $\forall$ i ($1 \leq i \leq n$), $\forall$ j ($1 \leq i \leq m$), $\Pi(\alpha_i)$ and $\Pi(\beta_j)$ are given. Also suppose that $\forall$ i, $\Pi(\alpha_i) = \Pi(\alpha_i | \beta)$, where $\beta$ is any formula that is built up from the $\beta_j$'s using the logical connectives and $\Pi(\beta) \neq 0$, and that $\forall$ j, $\Pi(\beta_j) = \Pi(\beta_j | \alpha)$, where $\alpha$ is any formula that is built up from the $\alpha_i$'s using the logical connectives and $\Pi(\alpha) \neq 0$. ("|" is Dempster's rule of conditioning.) Then $\Pi$ is that $\forall$ i, j, $\Pi(\alpha_i \wedge \beta_j) = \Pi(\alpha_i) * \Pi(\beta_j)$, where * is the usual multiplication operation.

**Proof:** First, we show that if there is a solution $\Pi$, then $\Pi$ must be such that $\forall$ i, j, $\Pi(\alpha_i \wedge \beta_j) = \Pi(\alpha_i)*\Pi(\beta_j)$. This is easy. $\forall$ j ($1 \leq j \leq m$), either $\Pi(\beta_j) = 0$ or $\Pi(\beta_j) \neq 0$. In the first case, $\forall$ i ($1 \leq i \leq n$), $\Pi(\alpha_i \wedge \beta_j) = 0 = \Pi(\alpha_i)*\Pi(\beta_j)$. In the second case, $\forall$ i ($1 \leq i \leq n$), $\Pi(\alpha_i) = \Pi(\alpha_i | \beta_j) = \Pi(\alpha_i \wedge \beta_j | \beta_j) = \Pi(\alpha_i \wedge \beta_j)/\Pi(\beta_j)$, that is, $\Pi(\alpha_i \wedge \beta_j) = \Pi(\alpha_i)*\Pi(\beta_j)$.

Next, we show that $\Pi$ is a solution, where $\Pi$ is such that $\forall$ i, j, $\Pi(\alpha_i \wedge \beta_j) = \Pi(\alpha_i)*\Pi(\beta_j)$. Without loss of generality, let us assume that $1 = \Pi(\alpha_1) > \Pi(\alpha_2) > ... > \Pi(\alpha_n)$ (just put those with the same possibility into one set). It suffices to just consider conditioning $\Pi$ on $\alpha$, where $\alpha$ is any formula that is built up from the $\alpha_i$'s using the logical connectives and $\Pi(\alpha) \neq 0$. $\{[\alpha_i \wedge \beta_j] : 1 \leq i \leq n, 1 \leq j \leq m\}$, a partition of $\Theta$, can be viewed as a matrix with n rows and m columns; each row represents an $\alpha_i$; each column represents a $\beta_j$; and each entry <i, j> represents the proposition $\alpha_i \wedge \beta_j$. [$\alpha$] then consists of several rows of this matrix. Let the ordinal numbers of these rows be $\mathcal{A}$ (e.g., if $\mathcal{A} = \{1, 3, 7\}$, then $[\alpha]$ consists of the first, third, and seventh row of the matrix; clearly, $\mathcal{A} \subseteq \{1, 2, ..., n\}$), and let k be the smallest element of $\mathcal{A}$. $\Pi(\alpha) = \Pi(\alpha_k)$. And so $\forall\, j\ (1 \le j \le m)$, $\Pi(\beta_j \mid \alpha) = \max\{\Pi(\alpha_i \wedge \beta_j \mid \alpha) : 1 \le i \le n\} = \max\{\Pi(\alpha_i \wedge \beta_j)/\Pi(\alpha) : i \in \mathcal{A}\} = \max\{\Pi(\alpha_i \wedge \beta_j)/\Pi(\alpha_k) : i \in \mathcal{A}\} = \max\{\Pi(\alpha_i)*\Pi(\beta_j)/\Pi(\alpha_k) : i \in \mathcal{A}\} = \max\{\Pi(\alpha_i) : i \in \mathcal{A}\}*\Pi(\beta_j)/\Pi(\alpha_k) = \Pi(\alpha_k)*\Pi(\beta_j)/\Pi(\alpha_k) = \Pi(\beta_j)$. □

---

[5]The only difference is that, when using their minimum rule, $\Pi(.|\alpha)$ is defined (to be 1) when $\Pi(\alpha) = 0$.

Clearly, Dempster's rule is the only one that is acceptable among the three rules that we considered. Also, it seems that no other natural normalization rule can be used in order that the notion of belief independence is supported (for example, it can be shown that Theorem 3.1 does not hold if the conditioning rule we use distributes the normalization constant evenly among the $c_i$'s). However, we cannot rule out all other normalization rules. Because we are talking about *all* "arbitrary but fixed rules" here. And so we can only say that Dempster's rule is the only natural variation of confidence transfer that is known to be "qualified" in supporting the notion of belief independence.

## 4 POSSIBILISTIC PROPAGATION

Suppose we do use Dempster's rule as our official rule of conditioning, then we can use the so-called "causal net" technique [Pearl 88] for modular specifications, and use local computation to compute conditional marginal possibilities from the joint possibility satisfying the given constraints. It turns out that the propagation scheme that we use is already described in Shenoy [92]. However, as we have pointed out in Section 1, our intuition is completely different from that of Shenoy [92]. For us, it is not just a local computation scheme that we want. There has to be a firm semantic basis underlying this computation scheme, so that we know how to use it (correctly) for reasoning. For this reason, we think it is worthwhile to describe our reasoning approach in its entirety, and not just part of it. This is what we do in this section. Below, we just give an informal illustration. The interested reader should consult [Pearl 88] for aspects of modular specifications and [Pearl 86; Lauritzen and Spiegelhalter 88; Shenoy 89, 92, 94; Shenoy and Shafer 90] for aspects of local computation.

The technique of modular specification starts with the specification of a directed acyclic graph $G$. Take Figure 4.1 as an example. It specifies what we think about the causal relations among a (possible) earth quake, a (possible) burglary into our house, the (possible) sound of an alarm in our house, (possible) radio announcements of the earth quake, and (possible) phone calls (informing us of the alarm) from our two neighbors Watson and Gibbon. Each node of $G$ represents a propositional primitive P, and contains two values T and F (representing two possible truth values of P). If $(P_1, Q), (P_2, Q), ..., (P_n, Q)$ are all the edges that point to Q in $G$, then it means that our judgment of the truth or falsity of Q is fixed *once the truth values of $P_1, P_2, ..., $ and $P_n$ are known* (independent of what the truth values of other primitives are). If a node P has no edge pointing to it (i.e., P is a "root node"), then it means that our judgment of P (a "prior") can be assessed directly without knowing the value of any primitive.

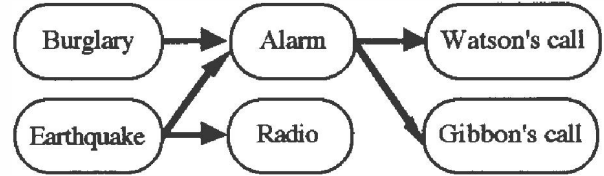

Figure 4.1: The alarm problem (from Pearl [88])

To specify a possibility measure $\Pi$, we just specify the priors and conditionals. That is, for each "root node" P, we specify $\Pi(P)$ and $\Pi(\neg P)$; and if $(P_1, Q), (P_2, Q), ..., (P_n, Q)$ are all the edges of $G$ that point to Q, then we specify $\Pi(Q|\alpha)$ and $\Pi(\neg Q|\alpha)$ for *every* formula $\alpha$ of the form $L_1 \wedge L_2 \wedge ... \wedge L_n$, where $L_i$ ($1 \le i \le n$) is either $P_i$ or $\neg P_i$. In terms of the above example, our specification can be $\Pi(B) = \Pi(\neg B) = \Pi(E) = \Pi(\neg E) = 1$. $\Pi(\neg A|B \wedge E) = .05$, $\Pi(\neg A|B \wedge \neg E) = .4$, $\Pi(\neg A|\neg B \wedge E) = .85$, $\Pi(A|\neg B \wedge \neg E) = .05$, $\Pi(\neg R|E) = .05$, $\Pi(R|\neg E) = 0$, $\Pi(\neg W|A) = .8$, $\Pi(W|\neg A) = \Pi(\neg W|\neg A) = 1$, $\Pi(\neg G|A) = .8$, and $\Pi(G|\neg A) = \Pi(\neg G|\neg A) = 1$. (We need not specify a value for, e.g., $\Pi(A|B \wedge E)$; its value has to be 1, because $\Pi(\neg A|B \wedge E)$ is less than 1.) Note that here, our priors for B and E just say that "everything is possible".

Given the above sample specification, we can compute the possibility of $L_B \wedge L_E \wedge L_A \wedge L_R \wedge L_W \wedge L_G$ ($L_B$ is either B or $\neg B$, $L_E$ is either E or $\neg E$, etc.) using a formula ((IV) below) that is derivable from the independence assumptions we made in specifying the graph.

$\Pi(L_B \wedge L_E \wedge L_A \wedge L_R \wedge L_W \wedge L_G)$

/* $= \Pi(L_B) * \Pi(L_E|L_B) * \Pi(L_A|L_B \wedge L_E)$

$* \Pi(L_R|L_B \wedge L_E \wedge L_A) * \Pi(L_W|L_B \wedge L_E \wedge L_A \wedge L_R)$

$\Pi(L_G|L_B \wedge L_E \wedge L_A \wedge L_R \wedge L_W)$ */

$= \Pi(L_B) * \Pi(L_E) * \Pi(L_A|L_B \wedge L_E) * \Pi(L_R|L_E)$

$* \Pi(L_W|L_A) * \Pi(L_G|L_A)$     (IV)

With formulas like (IV), we can compute, in theory at least, $\Pi(\{\omega\})$ for all $\omega$ (i.e., $L_B \wedge L_E \wedge L_A \wedge L_R \wedge L_W \wedge L_G$), while $\Pi(\beta|\alpha)$ is computed from $\Pi(\alpha \wedge \beta) / \Pi(\alpha) = \max\{\pi(\omega): \omega \in [\alpha \wedge \beta]\} / \max\{\pi(\omega): \omega \in [\alpha]\}$. But clearly, if there are many primitives, then it will not be feasible to directly compute $\Pi(\beta|\alpha)$ from the joint possibility. This is why local computation techniques are developed in the first place. Below, we describe a local computation scheme. This scheme computes, for each primitive P, $\Pi(P|\alpha)$ and $\Pi(\neg P|\alpha)$, where $\alpha$ is some conjunction of literals such as ¬W∧G∧R or ¬R∧W∧G.

Consider the alarm-problem again. There is a total of six specifications of Π-values. Each Π-specification involves one or more primitives, and the six groups of primitives are {B}, {E}, {B, E, A}, {E, R}, {A, W}, and {A, G}. It happens that we can arrange these six groups of primitives in a tree, with the property (called the *Markov property*) that if two nodes u and v contain the same primitive P, then P is in every node on the path between u and v. In general, it may be the case that the groups of primitives involved cannot be arranged in a tree with the Markov property. When that happens, some additional groups of primitives will have to be added [Shenoy and Shenoy 90]. We are also allowed to add new groups of primitives as desired, so long as the Markov property is preserved.

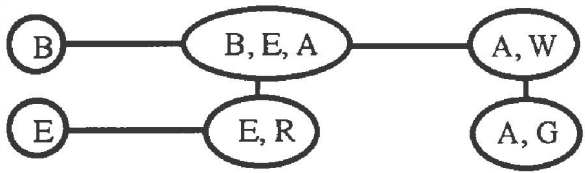

Figure 4.2: A *Markov tree*.

Originally, we specified some Π-values for each group of primitives. Now let us translate each of these Π-specifications into what is called a *potential*, and associate each potential with its corresponding node on the tree.

| B E A | | E R | A W | A G |
|---|---|---|---|---|
| T T T | 1.00 | T T 1.00 | T T 1.00 | T T 1.00 |
| T T F | 0.05 | T F 0.05 | T F 0.80 | T F 0.80 |
| T F T | 1.00 | F T 0.00 | F T 1.00 | F T 1.00 |
| T F F | 0.40 | F F 1.00 | F F 1.00 | F F 1.00 |
| F T T | 1.00 | | | |
| F T F | 0.85 | B | E | |
| F F T | 0.05 | T 1.00 | T 1.00 | |
| F F F | 1.00 | F 1.00 | F 1.00 | |

Figure 4.3: Potentials on the nodes of the Markov tree.

In this translation, Π(A|B∧E), for example, becomes the potential-value of the valuation (of {B, E, A}) B∧E∧A, while Π(¬B), for example, becomes the potential-value of the valuation ¬B. If there were any additionally introduced group of primitives, then each valuation of these primitives would be assigned a potential-value of 1.

We now need to give an informal introduction of the notion of *combination* and *marginalization*. Let g, h and k be the sets of primitives {B, E, A}, {B, E}, and {B, E, R}, respectively. If G is a potential on {B, E, A}, then $G^h$, the *marginal* of G for {B, E}, is such that $G^h(L_B \wedge L_E) = \max\{G(L_B \wedge L_E \wedge L_A)\}$ (we use $L_B$, for example, to mean either B or ¬B). *Combination* (of two potentials) is simply pointwise multiplication. That is, if we use G⊗K, where K is a potential on k, to denote the result of combining G and K, then G⊗K (a potential on g ∪ k) is such that $G \otimes K(L_B \wedge L_E \wedge L_A \wedge L_R) = G(L_B \wedge L_E \wedge L_A) * K(L_B \wedge L_E \wedge L_R)$. If there does not exist any $L_B \wedge L_E \wedge L_A \wedge L_R$ such that $G(L_B \wedge L_E \wedge L_A) * K(L_B \wedge L_E \wedge L_R) > 0$, then we say that G and K are not combinable.

Local computation is essentially a process of marginalizing and combining potentials in a node by node fashion. For illustrative purposes, we just describe a one-sweep process that (only) computes Π(B|W∧R) and Π(¬B|W∧R). First, we attach two additional nodes {W} and {R} to the Markov tree in Figure 4.2. Each of these two nodes is linked to a superset node (here we have no choice but to link {W} to {A, W}, and {R} to {E, R}), and is needed for input or output purposes. The potential on {W} is such that F is assigned 0 and T is assigned 1. The potential on {R} is the same. Let us designate the node {B} as the root of the tree. When the computation starts, each leaf just makes its potential "available" to its parent. Then, every internal node does the following: it waits until *all* its children have made their potentials available to it, and then it combines these potentials with its own potential, and makes the resulting potential available to its own parent if there is one. When the whole process finishes, Π(B|W∧R) and Π(¬B|W∧R) are just the potential on {B} normalized, i.e., we divide the potential-values of B and ¬B by their maximum.

Shenoy [92] has shown that the above described local computation scheme satisfies the three axioms of Shenoy and Shafer [90]. This, together with formulas such as (IV), imply that the scheme correctly computes the desired conditional marginals of the joint possibility satisfying the given constraints (i.e., the modular specification).

## 5  CONCLUSION

We have shown that the notion of confidence transfer can be justified from a logic-based perspective, and we do so by giving an axiomatization of confidence transfer. Our axiomatization is basically a strengthening of Rott's proposal [Rott 91], which was originally intended to be used in the area of theory of belief revision.

We also suggested that Dempster's rule of conditioning is the only *natural* variation of confidence transfer that is known to support the notion of belief independence. It remains to be shown, however, that Dempster's rule is the only variation of confidence transfer that supports the notion of belief independence (in the sense of Theorem 3.1).

One might question the intuitive legitimacy of our approach. After all, it is quite obvious that some other criteria could have been used in choosing among different proposals of conditioning rules. Why must we use the notion of belief independence? Moreover, our definition of conditional independence seems to have a strong probabilistic flavor. Is this a hint then that our version of possibility theory is perhaps just some transformation of probability theory? Our answer to these questions is as follows. To start with, we think that belief independence is an *a priori* notion. After all, there are things in the world that are independent of each other, conditionally or otherwise. As such, our opinions of these "things" ought to be independent. And so, as long as we take the notion of conditioning for granted, we should also take the notion of belief independence seriously. As for the definition of belief independence, it is just a matter of formulating our intuitive notion of belief independence in terms of our formal, generally accepted definition of conditioning. It is true that our definition of belief independence looks just like the one in probability theory. But this should be interpreted to mean that we are on the right track (is there any other way of formulating our intuitive notion of belief independence?), and not that our theory is some transformation of probability theory. It is true that the "formal structure" of our version of possibility theory resembles that of probability theory. But "similar" is by no means "equal", and it is important to note that these two theories have completely *opposite* notions of belief.

We also showed that if Dempster's rule is used as the "official" rule of conditioning, then we can use the so-called causal net technique to make modular specifications, and also use local computation to compute the desired conditional marginals of the underlying joint possibility. Our technique of modular specification is not new, nor is our technique of local computation. But we are the first to integrate these two techniques into possibility theory, using Dempster's rule of conditioning as the glue.

## Acknowledgements

The author has benefited from discussions with Robert Kennes, Prakash Shenoy, and Philippe Smets, and would also like to thank two anonymous referees for their helpful suggestions. This work was supported in part by the National Science Foundation of the Republic of China under Grant No. NSC 82-0113-E-033-054-T.

## References

Dubois, D. (1986). Belief structures, possibility measures and decomposable set-functions. *Computers and Artificial Intelligence*, Bratislava 5, 403-416.

Dubois, D. and Prade, H. (1986). Possibilistic inference under matrix form. *Fuzzy Logic in Knowledge Engineering* (H. Prade and C.V. Negoita, eds.), Verlag TÜV Rheinland, Köln, 112-126.

Dubois, D. and Prade, H. (1990a). Inference in possibilistic hypergraphs. In *Proceedings of IPMU-90*, Lecture Notes in Computer Science 521 (Bouchon-Meunier, B., Yager, R.R. and Zadeh, L.A. eds.) 250-259, Springer-Verlag.

Dubois, D. and Prade, H. (1990b). Epistemic entrenchment and possibilistic logic. *Artificial Intelligence* 50, 223-239.

Dubois, D. and Prade, H. (1991). Possibilistic logic, preference models, non-monotonicity and related issues. *Proceedings of IJCAI-91*, Sydney, Australia, 24-30.

Fonck, P. (1990). Building influence networks in the framework of possibility theory. *Proceedings of the RP2 First Workshop DRUMS*, Albi, France, 26-28.

Gärdenfors, P. and Makinson, D. (1994). Nonmonotonic Inference Based on Expectations. *Artificial Intelligence* 65 (2), 197-245.

Hisdal, E. (1978). Conditional possibilities - independence and non-interaction. *Fuzzy Sets and Systems*, 1, 283-297.

Hsia, Y.-T. (1991). Belief and surprise - a belief-function formulation. In *Proceedings of the Seventh Conference on Uncertainty in AI*, Los Angeles, California, 165-173.

Lauritzen, S. L. and Spiegelhalter, D. J. (1988). Local computations with probabilities on graphical structures and their application to expert systems. *Journal of the Royal Statistical Society*, series B 50 (2), 157-224.

Pearl, J. (1986). Fusion, propagation, and structuring in belief networks. *Artificial Intelligence* 29, 241-288.

Pearl, J. (1988). *Probabilistic Reasoning in Intelligent Systems: Networks of Plausible Inference*. Morgan Kaufmann Publishers, Inc., San Mateo, California.

Rott, H. (1991). Two methods of constructing contractions and revisions of knowledge systems. *Journal of Philosophical Logic* 20, 149-173.

Shackle, G. L. S. (1961). *Decision, Order and Time in Human Affairs*. Cambridge University Press.

Shafer, G. (1976). *A Mathematical Theory of Evidence*. Princeton University Press.

Shenoy, P. P. (1989). A valuation-based language for expert systems. *International Journal of Approximate Reasoning* 3, 383-411.

Shenoy, P. P. (1992). Using possibility theory in expert systems. *Fuzzy Sets and Systems* 52, 129-142.

Shenoy, P. P. (1994). Conditional independence in valuation-based systems. *International Journal of Approximate Reasoning* 10 (3), 203-234.

Shenoy, P. P. and Shafer, G. (1990). Axioms for probability and belief function propagation. *Proceedings of the Sixth Conference on Uncertainty in Artificial Intelligence*, Boston, Massachusetts, 169-198.

Smets, P. (1988). Belief functions. In *Non-Standard Logics for Automated Reasoning* (P. Smets, E. H. Mamdani, D. Dubois and H. Prade eds.). Academic Press, London.

Smets, P. (1993). Quantifying beliefs by belief functions : an axiomatic justification. In *Proceedings of IJCAI-93*, 598-603.

Yager, R. R. (1987). On the Dempster-Shafer framework and new combination rules. *Information Sciences* 41, 93-137.

Zadeh, L.A. (1978). Fuzzy sets as a basis for a theory of possibility. *Fuzzy Sets and Systems* 1, 3-28.